\documentclass{article}

\usepackage{PRIMEarxiv}

\usepackage[utf8]{inputenc} % allow utf-8 input
\usepackage[T1]{fontenc}    % use 8-bit T1 fonts
\usepackage{url}            % simple URL typesetting
\usepackage{booktabs}       % professional-quality tables
\usepackage{amsfonts}       % blackboard math symbols
\usepackage{amsmath}        % for \text and math environments
\usepackage{nicefrac}       % compact symbols for 1/2, etc.
\usepackage{microtype}      % microtypography
\usepackage{xcolor}         % color text
\usepackage{lipsum}
\usepackage{natbib}
\usepackage{fancyhdr}       % header
\usepackage{graphicx}       % graphics
\graphicspath{{media/}}     % organize your images and other figures under media/ folder
\usepackage{algorithm}
\usepackage{algorithmic} % for algorithmic environment
\usepackage{verbatim} % for comment environment
\usepackage{amsfonts} % for \mathbb
\usepackage{hyperref} % hyperlinks
\hypersetup{
    colorlinks=true,
    urlcolor=blue
}

\title{Flow-Based Generative Modeling for Optimizing Sampling Policies in Compressed Sensing Applications
}

\author{
  Roman Pavelkin, Luis A. Zavala-Mondragon, Christiaan G. A. Viviers, Fons van der Sommen \\
  Eindhoven University of Technology \\
  Eindhoven\\
  \texttt{\{r.pavelkin, l.a.zavala.mondragon, c.g.a.viviers, fvdsommen\}@tue.nl} \\
}

\begin{document}
\maketitle

\begin{abstract}

Numerous modern applications in signal processing and medical imaging necessitate acquiring high-dimensional signals under tight resource constraints. Traditional sampling theory suggests that accurate signal reconstruction requires a number of measurements proportional to the signal's ambient dimension, a requirement often too expensive or impractical. Compressed sensing challenges this notion by demonstrating that sparse signals can be recovered with fewer measurements, provided the measurement operator meets certain conditions. This proof-of-concept study presents a task-aware flow-based generative framework--a reformulation of the conventional Flow Matching training paradigm with a flow model trained to optimize subsampling in compressed sensing applications. We establish the fundamental feasibility of the proposed framework of learning subsampling masks that improve the performance of compressed sensing for image classification, image reconstruction, and MRI acceleration. For the image reconstruction task, our method achieved Peak Signal-to-Noise Ratio of 25.17 dB (improvement of 33\% over the SotA) at the subsampling rate of 5\% on the CelebA dataset and 29.24 dB of PSNR and 0.7163 of SSIM (improvement of 23\% over the SotA) when reconstructing $8\times$ accelerated MRI measurements (fastMRI dataset) with minimal computational overhead. These results highlight the effectiveness of task-conditioning within generative flow models and reveal a promising direction for representation learning strategies. Overall, the proposed framework offers a unified, flexible approach to designing data- and task-driven sensing schemes that can be potentially adapted to a broad range of inverse problems. The implementation is available on our \href{https://github.com/RomanPavelkin/flowcs}{GitHub repository}.

\end{abstract}

% keywords can be removed
% \keywords{First keyword \and Second keyword \and More}

\section{Introduction}
\label{intro}

Numerous contemporary applications in signal processing, medical imaging, and communications require the acquisition of high-dimensional signals under stringent resource constraints. Classical sampling theory asserts that accurate signal reconstruction necessitates a number of measurements proportional to the ambient dimension of the signal, a requirement that is frequently prohibitively expensive or physically infeasible. The theory of Compressed Sensing~(CS) challenges this traditional paradigm by establishing that sparse signals can be recovered from a substantially reduced number of measurements, provided that the measurement operator satisfies appropriate incoherence or restricted isometry properties~\cite{donoho_compressed_2006, candes_introduction_2008}. The associated reconstruction problem constitutes a specific instance of a more general class of Inverse Problems~(IP), in which one aims to infer unknown quantities from incomplete and indirect observations~\cite{engl_regularization_1996}. Such problems are intrinsically ill-posed in the absence of additional structural assumptions or regularity conditions imposed on the unknown. Moreover, it can be observed that in such settings, the quality of the final reconstruction is highly dependent on the choice of the sampling scheme~\cite{adcock_compressive_2021, manohar_data-driven_2018}. Consequently, finding an optimal subsampling pattern is critical. Nevertheless, this problem is intrinsically difficult: the sampling mask is discrete and non-differentiable, which prevents naive end-to-end optimization with gradient-based methods, and the search space itself is combinatorial, growing dramatically with the number of possible sensor locations. Exhaustive exploration is therefore computationally infeasible, while heuristic or hand-crafted sampling patterns often fail to align with the structure of the downstream reconstruction task. Consequently, generative-model-based methodologies have emerged, enabling the design of sampling schemes that can be tailored to both the specific data characteristics and the requirements of the downstream task~\cite{wu_deep_2019, huijben_deep_2019, van_gorp_active_2021, nolan_active_2024}. 

Generative diffusion models~\cite{yang_diffusion_2023} and their continuous-time counterparts based on Flow Matching~(FM)~\cite{lipman_flow_2022} have recently emerged as state-of-the-art tools for modeling complex data distributions, offering both strong expressiveness and stable training dynamics. These models learn to progressively transform noise into structured data by estimating local score or velocity fields that define a generative trajectory through the data manifold. Beyond high-fidelity image synthesis, an increasingly important application of diffusion- and flow-based generative models is the solution of IP~\cite{daras_survey_2024}. In this setting, the learned generative prior serves as a powerful regularizer that constrains reconstructions to lie on the manifold of realistic images, enabling performance that often surpasses traditional optimization- or CNN-based approaches. As a result, diffusion and flow matching methods have achieved state-of-the-art results across various inverse imaging tasks, including inpainting~\cite{yu_inpaint_2023, rout_theoretical_2023}, deblurring~\cite{chen_hierarchical_2023, kong_deblurdiff_2025}, super-resolution~\cite{wang_exploiting_2024}, and compressed sensing~\cite{chung_diffusion_2022, nolan_active_2024}.

This paper presents a proof-of-concept study focused on configuring the FM framework to output a sampling strategy optimized for a certain CS task and data samples. Consequently, the model operates on the manifold of the sampling space rather than explicitly within the signal space. This constitutes a comparatively novel paradigm, since diffusion- or FM-based models have predominantly been employed as regularizers in inverse problem solvers within frameworks such as Plug-and-Play (PnP), Regularization by Denoising (RED), or Score-Based Posterior Sampling~\cite{daras_survey_2024}. However, training a FM model within the manifold of the sampling space offers several concrete advantages. The following can be highlighted:
\begin{itemize}
    \item The manifold associated with the sampling space is intrinsically subject to stronger constraints than the manifold of natural images. Leveraging these structural properties—such as binarity and sparsity—creates opportunities to improve the efficiency and effectiveness of model training.
    \item It provides continuous evolution of sampling scheme rather than a static mapping, as in, for example, DPS~\cite{huijben_deep_2019}, thereby ensuring a smooth, time-continuous parametrization of the underlying manifold.
    \item The time step $t$ acts as an implicit entropy-regularization term in both the generative pathway and the gating mechanism employed for the construction of soft masks~(binary mask relaxation).
    \item It allows for joint optimization with a downstream CS task, and therefore benefits from the task prior.
    \item The framework naturally handles uncertainty in sampling strategies as the model outputs a distribution
\end{itemize}
To the best of the authors’ knowledge, such a task-tailored reconceptualization of flow matching has not previously been documented in the literature. The closest analogy can be found in the work~\cite{kaleta_jointdiffusion_2025}, which introduced JointDiffusion. JointDiffusion is a diffusion model jointly trained with a classifying network for simultaneous data generation and classification. It demonstrated how combining the two tasks helps improve model performance in both cases.

As a result, we establish the fundamental feasibility of the flow-based CS-tailored framework proposed in this paper to generate a compressed sensing policy that maximizes a specific downstream CS task. Specifically, the CS tasks studied are compressed multiclass classification, image restoration from compressed measurement, and accelerated Magnetic Resonance Imaging~(MRI) reconstruction. The latter case study is likely the most clinically meaningful, since scan time is a major limiting factor in practice~\cite{lustig_compressed_2008}. Long acquisitions heighten patient discomfort and motion artifacts while also reducing scanner throughput and limiting access to high‑resolution diagnostic data.

The architecture of the flow models has been implemented with U-Net~\cite{ronneberger_u-net_2015} or multilayer perceptron~(MLP) backbones, depending on the specific requirements for each task. Experiments have been conducted on public datasets (MNIST~\cite{lecun_mnist_1998}, CelebA~\cite{liu_deep_2015}, and fastMRI~\cite{zbontar_fastmri_2018}). We show that the proposed approach excels on compressed image and MRI reconstruction, where it surpasses the current best-performing and state-of-the-art methods, particularly at more aggressive sampling rates. Moreover, our method performs on par, or close to state-of-the-art approaches for the remaining tasks. 

This manuscript is organized as follows. Section~\ref{methods} formally introduces the CS problem and provides a concise overview of state-of-the-art solution strategies based on diffusion models. This section also describes our original methodological contribution. Section~\ref{res_disc} presents a comparative empirical evaluation of our model against established baselines across a range of experimental settings. In Section~\ref{conc}, we outline prospective research directions and potential application domains for the proposed approach. Finally, we provide additional experimental details in Appendices.

\section{Methods}
\label{methods}

\subsection{Signal and task-adaptive models in compressed sensing}

The problem of reconstruction of subsampled signals can be formalized as an inverse problem~(IP) that recovers an unknown signal $s \in \mathbb{R}^N$ from incomplete observations $z \in \mathbb{R}^M$, with $M < N$. The samples are obtained by applying the sampling operator (measurement matrix) $A\in \{0,1\}^N$ to the signal $s$. In imaging applications, $A$ typically corresponds to a masking operator in the spatial or transform domain, yielding the forward model:
\begin{equation}
    z = As + n, 
    \label{eq:eq_1}
\end{equation}
where $n$ denotes measurement noise. It should be noted that $A$ is not invertible. Therefore, the inverse problem of recovering $s$ is ill-posed, admitting infinitely many solutions consistent with the observations, and therefore requires the incorporation of prior information to constrain the solution~space~\cite{donoho_compressed_2006, geethanath_compressed_2013}. 

Deep neural networks are increasingly used as alternatives to traditional compressed sensing solvers, providing data-driven priors and quick inference. One major approach, algorithm unrolling, uses fixed-depth networks aligned with iterations of classical algorithms like ISTA or ADMM~\cite{gregor_learning_2010}, leading to efficient parameter learning from training data, as seen in models like MoDL~\cite{aggarwal_modl_2018} and E2E-VarNet~\cite{sriram_end--end_2020}. Another approach, the Plug-and-Play (PnP) framework, utilizes pretrained denoising networks to replace proximal operators, enabling flexible use of various denoisers without needing explicit regulariser formulations~\cite{ryu_plug-and-play_2019}. Furthermore, score-based diffusion models have been integrated into PnP frameworks, improving performance in MRI and other imaging tasks by learning data distribution scores and melding them with measurement likelihood in a reverse diffusion process, exemplified by diffusion posterior sampling and Pseudoinverse-Guided Diffusion Models ($\Pi$GDM)~\cite{song_pseudoinverse-guided_2023}. These methodologies efficiently maintain data fidelity while leveraging learned priors without the need for retraining for different measurement operators.

We now shift our attention from the CS reconstruction algorithm to methods that optimize the sampling operator $A$ for a particular task function $f$ and/or specific datasets. Since estimation of such a solution is the NP-hard combinatorial problem of finding an optimal subgroup within the group, it is often relaxed to the objective of learning the distribution of all possible operators $A_\phi$ parametrized by $\phi$:
\begin{equation}
    A_\phi \sim P(A \mid \phi).
    \label{eq:eq_2}
\end{equation}
Consequently, instead of directly optimizing $A_\phi$, we optimize the distribution’s parameters $\phi$ themselves.

Bahadir~et~al.~\cite{bahadir_learning-based_2019} formulated the sampling problem by learning pixel-based thresholding of samples drawn from a uniform distribution, referred to as Learning-based Optimization of the Undersampling PattErn~(LOUPE). The paper proposes an end-to-end, data-driven method for jointly learning the $k$-space subsampling pattern and the corresponding reconstruction model under a fixed sparsity constraint. By using retrospectively undersampled data, the method customizes both the sampling pattern and a modified U-Net architecture (augmented with a forward undersampling model) to the specific image distribution in the training set. A clear limitation of this algorithm is that it employs a fixed mask, thereby reducing its adaptability and constraining its capacity to accommodate varying conditions or inputs.

Huijben~et~al.~\cite{huijben_deep_2019} adopted a probabilistic perspective in the Deep Probabilistic Subsampling~(DPS) framework. Instead of directly learning a specific subsampling scheme, they formulate an optimization problem over a generative model that draws samples from a probability distribution that expresses belief over effective subsampling patterns, and consequently optimize the parameters of this distribution. To enable differentiable sampling from a categorical distribution, Gumbel-softmax sampling was employed, offering a continuous relaxation of the sampling process~\cite{jang_categorical_2016, maddison_concrete_2016}. This facilitates joint end-to-end optimization of the subsampling strategy and the downstream model, thereby extending DPS to applications that go beyond merely reconstructing from subsampled data. Similar to the LOUPE algorithm, DPS also produces a fixed subsampling mask; however, it achieves superior performance compared to LOUPE.

The modification of DPS presented by Van~Gorp~et~al.~\cite{van_gorp_active_2021}, known as Active DPS (A-DPS), offers \emph{sequential} adaptive sensing. It is achieved by introducing dependency between samples via a context vector produced by a Long Short-Term Memory~(LSTM) model that encodes information about the current task. In this setting, the parameters of the LSTM model are optimized instead of the direct optimization, as is done in DPS. Although this framework enables the specification of a temporal hierarchy of sampling points, its applicability to real-world scenarios remains limited (for instance, it is trained to operate at a single, fixed sampling rate, and given that A-DPS achieves marginal performance gain as compared to DPS~\cite{van_gorp_active_2021}). Furthermore, it continues to generate the same fixed subsampling mask as the standard DPS method.

A recent method introduced by Nolan~et~al.~\cite{nolan_active_2024}, termed Active Diffusion Subsampling~(ADS), outperforms LOUPE, DPS, and A-DPS across a range of undersampled reconstruction tasks. It leverages the ability of generative diffusion models to generate belief distributions. The framework relies on an accurate model of the posterior distribution over future measurements, given the measurements observed so far. The sample selection policy is based on maximum-entropy sampling across the posterior samples generated with the previously acquired signal points. This method demonstrates robustness across several signal domains and requires no additional task-specific training. However, the primary limitation of this framework is its substantial computational runtime, which significantly constrains the practical deployment of ADS in applications with stringent low-latency requirements, such as accelerated MRI or ultrasound scan-line subsampling.

\subsection{Conventional Flow Matching}

For self-containment, the current section presents a basic introduction to FM models. If more in-depth information is needed, we refer the reader to the works of Lipman and Lai~\cite{lipman_flow_2022, lai_principles_2025}.

\begin{figure}[tb]
    \centering
    \includegraphics[width=1.0\linewidth]{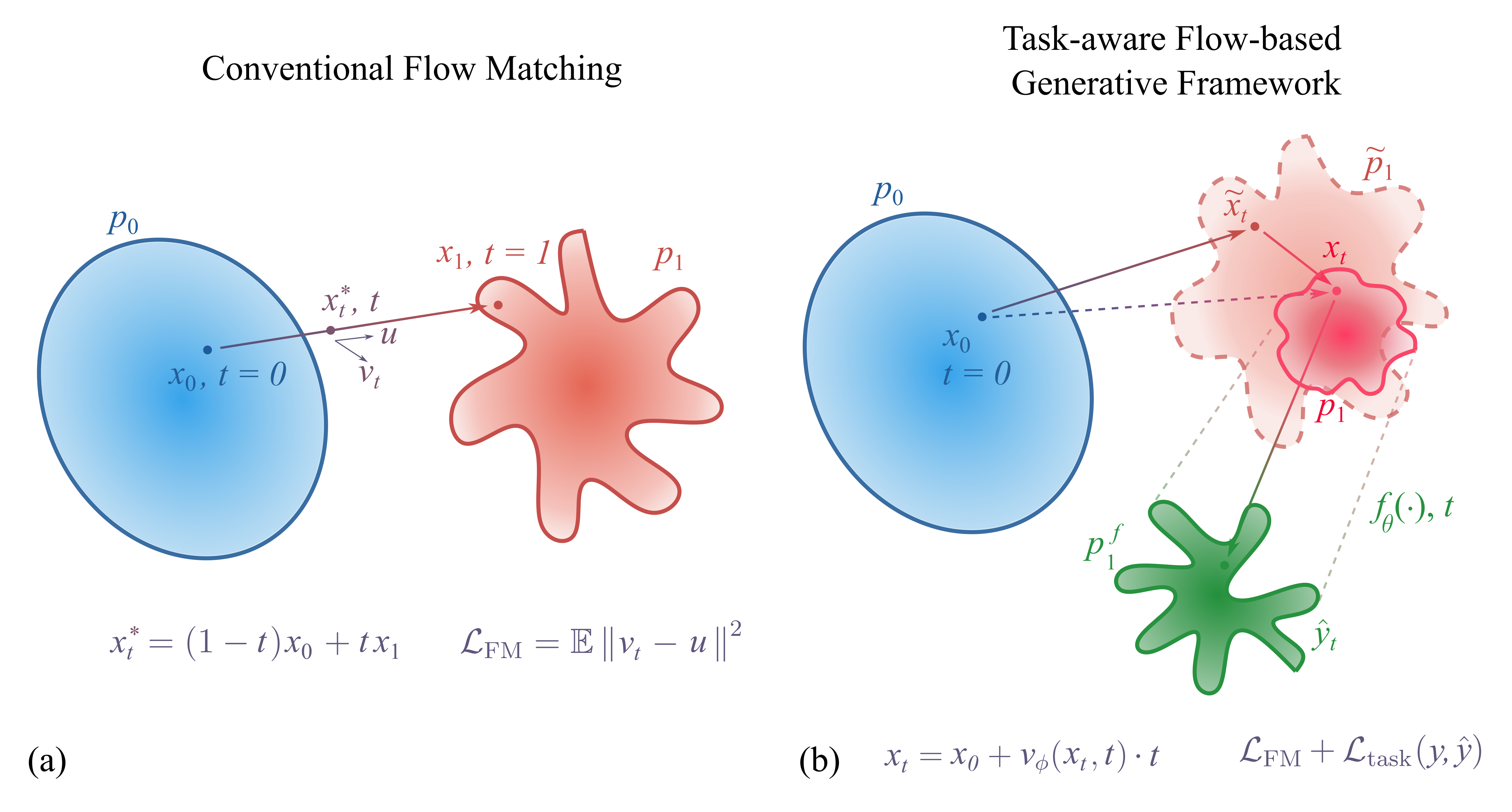}
    \caption{Training workflows of the (a) conventional flow matching set-up and (b) proposed task-aware flow-based framework. In conventional FM the model $v$ is trained to move samples from noise $p_0$ to a data distribution $p_1$ explicitly (see Eq.~\ref{eq:eq_4}-\ref{eq:eq_5}), having access to the ground truth (training data) $x_1$. This configuration corresponds to a scenario in which the complete signal is reconstructed directly from the noise. In contrast, our framework is trained to generate $x_t$ (sampling operator) indirectly, being guided by a CS task/solver $f_\theta$ via end-to-end backpropagation (see Eq.~\ref{eq:eq_11}). $v$ operates in sampling operator space $A_\phi$ (Eq.~\ref{eq:eq_2}), whether $f_\theta$ operates in signal space.}
    \label{fig:fig_1}
\end{figure}

Being a distribution matching framework, FM learns a time-dependent vector field $v(x, t)$ for a neural ODE 
\begin{equation}
    \dot{x} = v(x,t)
    \label{eq:eq_3}
\end{equation}
that deterministically transports points from the standard Gaussian distribution (noise) $p_0$ to the target data distribution $p_1$ (see Figure~\ref{fig:fig_1}(a)). The learning objective in this framework is to minimize the $L^2$ regression loss between the learned velocity field $v(x, t)$ and the ground truth velocity $u$, which defines the actual movement of the point from $p_0$ to $p_1$:
\begin{equation}
    \mathcal{L}_{\text{FM}} = \mathbb{E}\left\| v(x, t) - u \right\|^2,
    \label{eq:eq_4}
\end{equation}
where $u$ is typically expressed through the linear coupling assumption between $p_0$ and $p_1$. Thus, at a time step $t$, the ground truth path $x^{\star}_t$ is defined as
\begin{equation}
    x^{\star}_t = (1 - t)\,x_0 + t\,x_1, \qquad t \in [0,1],
    \label{eq:eq_5}
\end{equation}
where $x_0 \in p_0$ and $x_1 \in p_1$. As the formulation~Eq.~(\ref{eq:eq_5}) defines a straight line, each pair $(x_0, x_1)$ has a constant ground-truth velocity $u = \frac{d}{dt} x_t = x_1 - x_0$ along the~path~$x_t$. 

Once a network is trained to output $v$ given $t$ and $x_t$, the generated $\hat{x}$ is obtained via the forward Euler method integrating Eq.~(\ref{eq:eq_3}) over time:
\begin{equation}
    \hat{x} = x_0 + \int_{0}^{1} v(x, t) \, \text{d}t.
    \label{eq:eq_6}
\end{equation}
The sampling process fundamentally relies on the step count of the selected ODE solver, influencing its speed. During training, focus is often placed on ensuring a smooth transition of $v$ to maintain the ODE's well-conditioned nature, marked by a low Lipschitz constant for stable numerical integration~\cite{lai_principles_2025}. Additionally, it is possible to integrate multiple samples concurrently, as each sample independently follows the same ODE field.

\subsection{Task-aware adaptation of flow matching to learn the distribution of sampling operators}

In this section, we present our original contribution, which directly targets the optimization problem defined in~\ref{eq:eq_2}. To this end, we reformulate the conventional FM as an end-to-end trainable model by embedding a CS solver within the training pipeline. We then investigate the structure and properties of the resulting composite loss function. Furthermore, we introduce a sigmoid-based relaxation of the binary sampling operators into the training procedure of our generative framework and describe the details of this approach below.

\subsubsection{Coupling Flow Matching with a downstream task objective}

Let us consider a setting in which the ground-truth distribution \(p_1\) is not directly accessible and only an empirical or approximated estimate \(\tilde{p}_1\) can be obtained. At the same time, we posit the hypothesis that the true distribution either coincides with the estimated one or is contained within the family of distributions represented by \(\tilde{p}_1\):
\begin{equation}
    \text{supp}(p_1) \subseteq \text{supp}(\tilde{p}_1),
    \label{eq:eq_7}
\end{equation}
or overlaps with it:
\begin{equation}
    \text{supp}(p_1) \cap \text{supp}(\tilde{p}_1) \neq \emptyset.
    \label{eq:eq_8}
\end{equation}
Continuing to consider the kinematic particle motion defined by~Eq.~(\ref{eq:eq_3}), the marginal path $x_t$ of the points from noise $p_0$ to the proximal distribution $\tilde{p}_1$ at any time $t$ is defined as
\begin{equation}
    x_{t} = x_0 + v(x_t, t) \cdot t.
    \label{eq:eq_9}
\end{equation}

For the sake of clarity, it is now necessary to establish an analogy between the proximal distribution $\tilde{p}_1$ and the distribution of sampling operators (see Eq.~(\ref{eq:eq_2})). 
\begin{equation}
    A_\phi \sim \tilde{p}_1(x_t \mid \phi).
    \label{eq:eq_10}
\end{equation}
Therefore, the samples $x_t$ will be further denoted as CS sampling operators. A downstream CS task solver is further denoted as $f_\theta$. Thus, the flow network $v_\phi$ parametrized by $\phi$ transports the points from noise to the space of sampling operators $p_1$ via Eq.~(\ref{eq:eq_9}), whether the function $f_\theta$ parametrized by $\theta$ solves a CS task given the sampling operator $x_t$ on its input. Once both the flow $v_\phi$ and the task $f_\theta$ models are differentiable, learning the transformation dynamics $x_0 \to x_1$ becomes an end-to-end optimization problem driven by the downstream loss $\mathcal{L}_{\text{task}}(y, \hat{y})$:
\begin{equation}
    \frac{\partial \mathcal{L}}{\partial \phi} = \frac{\partial \mathcal{L}}{\partial \hat{y}} \cdot \frac{\partial \hat{y}}{\partial x_t} \cdot \frac{\partial x_t}{\partial \phi}.
    \label{eq:eq_11}
\end{equation}
Figure~\ref{fig:fig_1}~(b) displays a diagram of this generative framework.

It can be observed that the time weighting in Eq.~(\ref{eq:eq_9}) acts as an implicit regularizer: early times $(t \to 0)$ correspond to coarse, data- and task-agnostic structure, while later times $(t \to 1)$ capture task-specific refinement. The task $f_\theta$ anchors the flow's endpoint in a meaningful objective space serving as a prior. At the inference, the sampling policy $\hat{x}_1$ is generated with Eq.~(\ref{eq:eq_6}) via the forward Euler method.

As we further illustrate in the experiments, the proposed task-aware, flow-based generative modeling of a sampling policy that optimizes a task fits naturally within the IP paradigm, where it is understood that IPs do not have a single correct solution but instead yield estimates of an unobservable underlying reality.

\subsubsection{Composite training objective}

In practice, when training the entire generative framework given the optimization scope~(Eq.~(\ref{eq:eq_11})), it was found that the best results are achieved when combining the classical FM loss $\mathcal{L}_{\text{FM}}$ with a task-specific objective $\mathcal{L}_{\text{task}}$. Consequently, the total loss becomes:
\begin{equation}
    \mathcal{L}_{\text{total}}(\phi,\sigma_{1},\sigma_{2})=\frac{1}{2\sigma_1^{2}}\mathcal{L}_{\text{FM}} + \frac{1}{2\sigma_2^{2}}\mathcal{L}_{\text{task}} + \log\sigma_{1}\sigma_{2},
    \label{eq:eq_12}
\end{equation}
where $\sigma_{1,2}$ - the weighting parameters originating from the models' observation noise (how much noise we have in the outputs~\cite{kendall_multi-task_2018}).

The uncertainty-based weighting for the two loss components~(Eq.~(\ref{eq:eq_12})) results in optimizing the weights of those components jointly with the flow model's parameters $\phi$. Therefore, $\mathcal{L}_{\text{total}}$ becomes effectively “self-tuning”, balancing the model training process on complex composite tasks. The balancing is realized between enforcing global consistency of the generated sampling operators and the estimated ground-truth distribution (flow $p_0 \rightarrow \tilde{p}_1$), and subsequently refinement $\tilde{p}_1 \rightarrow\ p_1$ via $f_\theta$. Additional information regarding the approach used to obtain an estimate of the ground-truth distribution $\tilde{p}_1$ can be found~in~Appendix~\ref{append_a}.

\subsubsection{Differentiable Sigmoid Gate}

Given a generative flow model $v_\phi$, a downstream task $f_\theta$, and a specific dataset, we are interested in learning the optimal subsampling scheme conditioned both on the dataset and the task. However, the ground truth subsampling masks belong to binary space, which poses the fundamental limitation on backpropagation through step-like functions while training the model. To circumvent the non-differential nature of discrete sampling, the output $x_t \in \mathbb{R}^N$ from $v_\phi$ undergoes the following sigmoid transformation:
\begin{equation}
    g(x_t) = \mathcal{S}\!\left( s \cdot (x_t - \tau) \right), \quad s=30t
    \label{eq:eq_13}
\end{equation}
where $\mathcal{S}(\cdot)$ is the sigmoid function, $\tau = \mathrm{Quantile}_p(x_t)$ is the percentile threshold for soft-selection of the top $M$ entities in $x_t$, $s$ is the heuristically defined time-dependent steepness function, $t$ is the time step. The output of Eq.~(\ref{eq:eq_13}) $g(x_t) \in (0,1)$ represents a soft, differentiable sampling mask for the task: elements of $x_t$ above the chosen percentile $M$ receive values close to $1$, while elements below are smoothly suppressed toward $0$. The differentiable given in Eq.~(\ref{eq:eq_13}) is analogous to the differentiable Top-K operator~\cite{zhu_differentiable_2025}. However, our implementation includes $s(t)$, which serves as a way to regularize the steepness of $\mathcal{S}(\cdot)$: increasing $t$ makes the transition sharper, approaching hard thresholding. The aforementioned features of this custom sigmoid gating make it more preferable in our task as compared to, for example, the straight-through estimators~\cite{yin_understanding_2019}, which are the popular alternative for relaxation of discrete signals~\cite{huijben_review_2022}.

Appendix~\ref{append_a} shows that the sigmoid garing operator also can have an entropic interpretation of $t$. Moreover, Appendix~\ref{append_a} also shows an ablation study on the parameters of $s(t)$ function. The gated output $g(x_t)$ is then passed to $f_\theta$. % At the inference, the latent $\hat{x}_1$ is transformed with Eq.~(\ref{eq:eq_9}) for $t = 1$.

\subsection{Experiments}

In order to show the applicability of the proposed task-aware flow-based generative CS sampling framework, this paper explores two distinct CS tasks: (1)~classification of images obtained with compressed measurements and (2)~image reconstruction from compressed measurements. The investigation involves subsampling in pixel-space as well as in $k$-space. The latter is pertinent for frequency domain data acquisition scenarios, including but not limited to magnetic resonance imaging.

\begin{figure}[tb]
    \centering
    \includegraphics[width=1.0\linewidth]{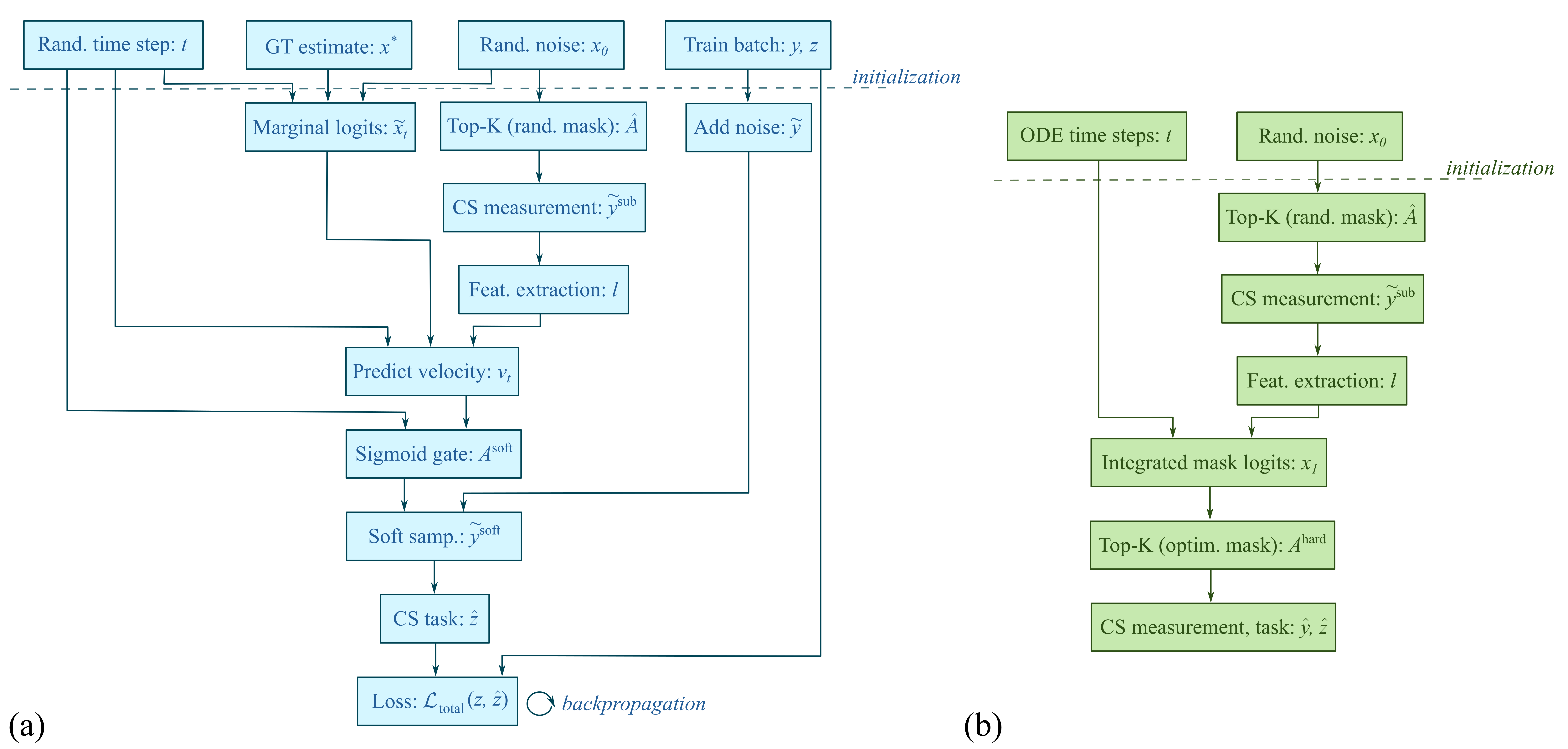}
    \caption{Workflow of (a)~training and (b)~inference of the flow-based generative sampling policy model. Diagram~(a) illustrates the Algorithm~\ref{alg:tagm_algorithm} from Appendix~\ref{append_b}.}
    \label{fig:fig_2}
\end{figure}

Our framework was examined in two modes. In the first mode, it is assumed that the signal is first measured at random points. This measurement is passed through a network trained to extract features from randomly sampled measurements. These features guide the flow model, which subsequently generates a data- and task-tailored measurement scheme. The illustration of training and the operational principle of the proposed framework at the inference is given in Figure~\ref{fig:fig_2}. 
This operating mode is appropriate when the signal varies only slightly between frames or when measurements are inexpensive enough to be performed twice: initially at random and subsequently in an optimized manner. The framework was tested in this regime for the experiments on MNIST (compressed classification and reconstruction) and CelebA (compressed reconstruction) datasets.

In MRI acceleration experiments, a subset of the Low-Frequency~(LF) bands in the k-space is first acquired. The MRI image reconstructed from the LF data is then processed by a feature-extraction model. The resulting features are used to condition a flow model, which determines the optimal sampling pattern for the remaining high-frequency k-space bands, where additional measurements are collected. The final, adaptively sampled \textit{k}-space data is subsequently processed by a task model. Consequently, this regime is particularly appropriate for scenarios in which each individual measurement is costly, making it highly suitable for MRI acquisition.

In each of the experiments reported below, the flow model $v_\phi$ was trained on a frozen task $f_\theta$; at inference, the ODE solver was executed for 20 time steps. The pseudocode of the algorithm for learning the subsampling mask optimized for a CS task is listed in Appendix~\ref{append_b}.

\subsubsection{Compressed classification of MNIST digits}
\label{methods_class_mnist}

The compressed classification framework was built for the MNIST dataset~\cite{lecun_mnist_1998} and tested at different subsampling rates (ratio $M/N$), ranging from 8\% to 1\%. For this task, the U-Net architecture~\cite{ronneberger_u-net_2015} was selected as the backbone for the flow model $v_\phi$. The design hyperparameters for this model are shown in Table~\ref{tab:tab_3} in Appendix~\ref{append_c}. To enable task-aware conditional generation, the reconstruction model $\mathcal{R}^{(L)}$ is pre-trained and used as a frozen feature extractor. The flow model $v_\phi$ consists of two parallel encoder pathways: one processes the conditioning input $x_{\mathrm{cond}}$ through $\mathcal{R}^{(L)}$, while the other processes noise $x_0$ via a trainable encoder. Multiscale features are fused across these pathways at each hierarchical level through convolutional projections, with time embeddings integrated at the bottleneck. The task loss term was the categorical cross-entropy. During training, $\mathcal{R}^{(L)}$ remains frozen, serving solely to provide multiscale latent representations that guide the flow matching process. Comprehensive architectural specifications and training hyperparameters are detailed in Appendix~\ref{append_c}.

\subsubsection{Compressed image reconstruction}
\label{methods_recon_mnist}
Experiments for this section follow the guidelines of the experiment with the compressed classification of MNIST digits. In order to make a proper comparison with the state-of-the-art model ADS from~\cite{nolan_active_2024} designed for undersampled reconstruction of the MNIST digits of the resolution 32~$\times$~32 pixels, we upscale the dataset accordingly. The number of subsampling pixels varies from 10 to 500 pixels during training and inference. The task model for this application mirrors the encoder $\mathcal{R}^{(L)}$ and, in fact, is the same model, which was pre-trained for feature extraction and constraining the generative flow model $v_\phi$ during its training.

To replicate the experiment of Nolan~et~al.~\cite{nolan_active_2024} with natural image reconstruction with the CelebA dataset, the network $\mathcal{R}$ has been adjusted to handle three-channel RGB images from the CelebA dataset with resolution 128~$\times$~128 pixels. A test set of 100 images was selected.

\subsubsection{MRI acceleration}
\label{methods_mri_acc}
To demonstrate the real-world practicability of the proposed generative framework, it was configured for the task of MRI acceleration. In this task, one reconstructs a fully-sampled MRI image given a budget of only a certain percentage of the $k$-space measurements, where each $k$-space measurement is a vertical line of width 1 pixel. The popular knee fastMRI dataset~\cite{zbontar_fastmri_2018} was chosen for the experiments. For simplicity, we consider only the single-coil measurements.

Several studies in the literature have proposed subsampling algorithms for accelerating MRI utilizing the fastMRI dataset, achieving state-of-the-art performance. To ensure a fair comparison with their results, we conducted two series of experiments on MRI slices of different resolutions and with different acceleration factors, namely: slices of 208$\times$208 pixels with 8$\times$ acceleration as done by  Van~Gorp~et~al.~\cite{van_gorp_active_2021} and slices of 128~$\times$128 pixels with $4\times$ acceleration~\cite{nolan_active_2024}.

Files in the dataset represent volumes of complex-valued images. From each volumetric scan, slices in proximity to the central slice, specifically within a range of ±10 slices, have been extracted with a patient-level split of the dataset. This way, we obtained the training set with 20,433 slices and the test set with 4,179 slices. The selected slices were cropped around their central region to achieve the specified resolution. 

As the task model $f_\theta$ for MRI reconstruction from the subsampled measurements, the notable Model-Based Deep Learning~(MoDL) framework proposed in the work by Aggarwal~et~al.~\cite{aggarwal_modl_2018} was employed. A comprehensive description of our MoDL implementation is provided in Appendix~\ref{append_c}.

It can be noted that in the current setting, it is sufficient to sample along one axis, as the mask for $k$-space represents a set of vertical lines. Thus, the architecture of the flow model has been defined as MLP (see Table~\ref{tab:tab_5} in Appendix~\ref{append_c} for details).

\section{Results and discussions}
\label{res_disc}
\subsection{Classification of MNIST digits}
\label{res_class_mnist}
\begin{figure}[tb]
    \centering
    \vspace{0.1in}
    \includegraphics[width=0.5\linewidth]{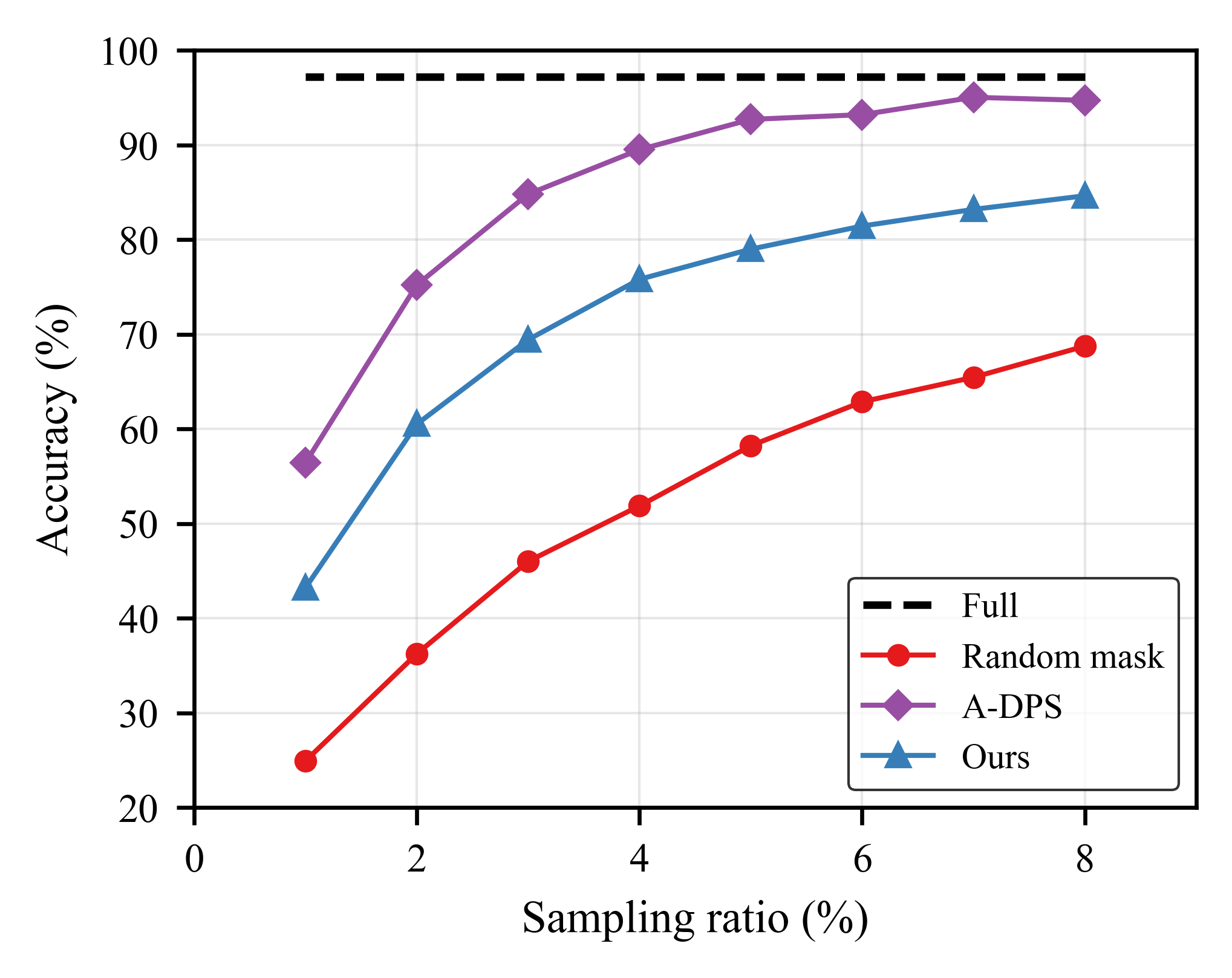}
    \caption{Classification accuracy for MNIST over different subsampling ratios. The \textit{full} line indicates the accuracy achieved for the classifying network trained without subsampling.}
    \label{fig:fig_3}
\end{figure}

The results of the digit classification accuracy for the task-aware flow-based framework~(Figure~\ref{fig:fig_3}) clearly demonstrate that the model learns sampling operators that improve the accuracy by more than 20\%, when compared to a classifier trained on randomly-generated masks. However, flow-generated sampling schemes for MNIST classification are less accurate than those provided by the A-DPS model~\cite{van_gorp_active_2021}. The difference in accuracy is 14\% for sub-sampling rates between 6\% and 8\%, and the gap increases as the number of sampling pixels decreases. We hypothesize that the existing performance gap between our algorithm and A-DPS may be caused by the fact that the A-DPS framework independently trains for each specified sampling ratio, whereas our algorithm was trained over a range of ratios from 1\% to 8\%, while our method exhibits greater versatility across different sampling ratios, at the cost of a reduction in accuracy. Note that concurrently with A-DPS, the flow-based generative framework inherently supports active sensing, while eliminating the requirement to tailor or overfit the model to specific sampling ratios.

\subsection{Image reconstruction}
\begin{figure}[tb]
    \centering
    \includegraphics[width=1.0\linewidth]{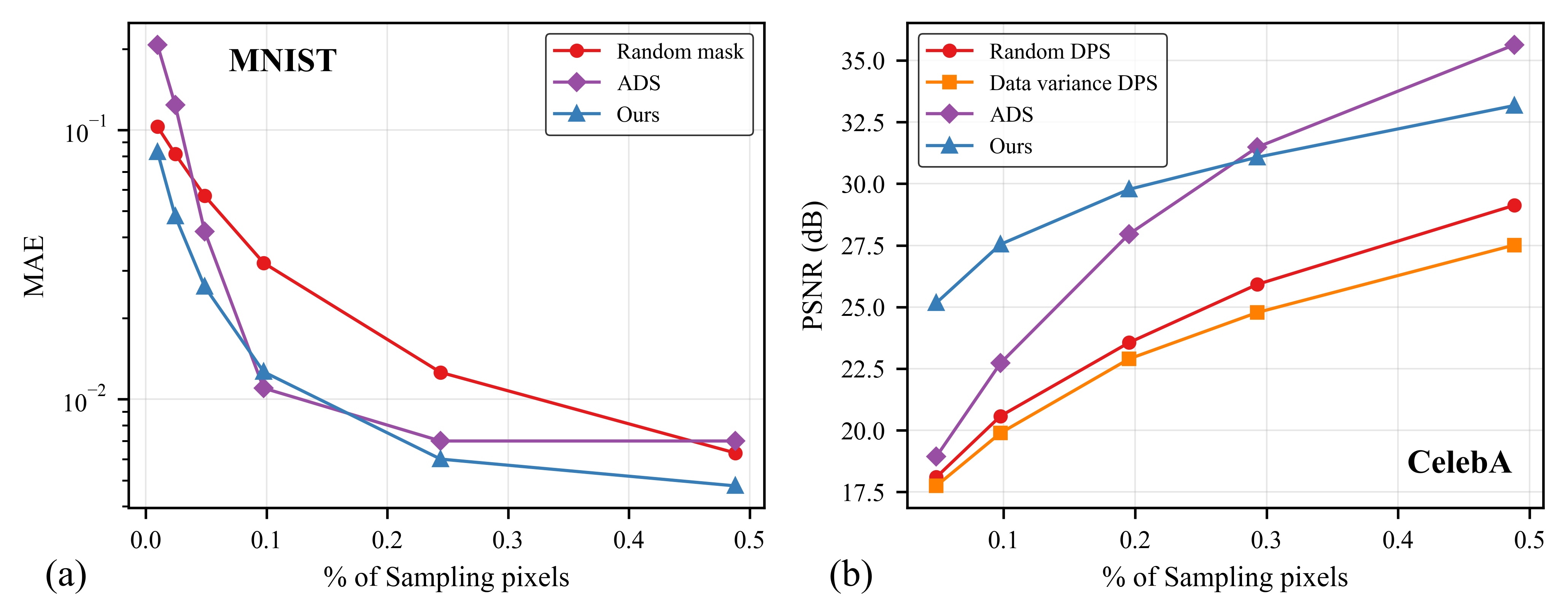}
    \caption{Image reconstruction metrics over different subsampling rates: (a) Mean absolute error (MAE) for MNIST digits; (b) PSNR for natural CelebA images.}
    \label{fig:fig_4}
\end{figure}

In the case of CS image reconstruction, the flow-based framework shows clear superiority both over the U-Net model trained for reconstruction with random sampling masks and over the baseline~(ADS) for both, the MNIST adn CelebA datasets~(see Figure~\ref{fig:fig_4}). Note that our method achieves superior reconstruction than ADS for lower sampling ratios, while ADS improves its performance for higher sampling rates.

For instance, the ADS method exhibited higher Mean Absolute Error~(MAE) for MNIST digit reconstruction, when compared to the flow-based algorithm, particularly for masks containing 1\%, 2\%, 5\%, and 49\% of sampling pixels. The sampling schemes generated with flow for the reconstruction of the CelebA images provide the highest PSNR values across subsampling rates of 5\%, 10\%, and 20\% pixels. At the least aggressive sampling rates, ADS demonstrates comparable or better reconstruction for the generated masks.

\begin{figure}[tb]
    \centering
    \includegraphics[width=0.5\linewidth]{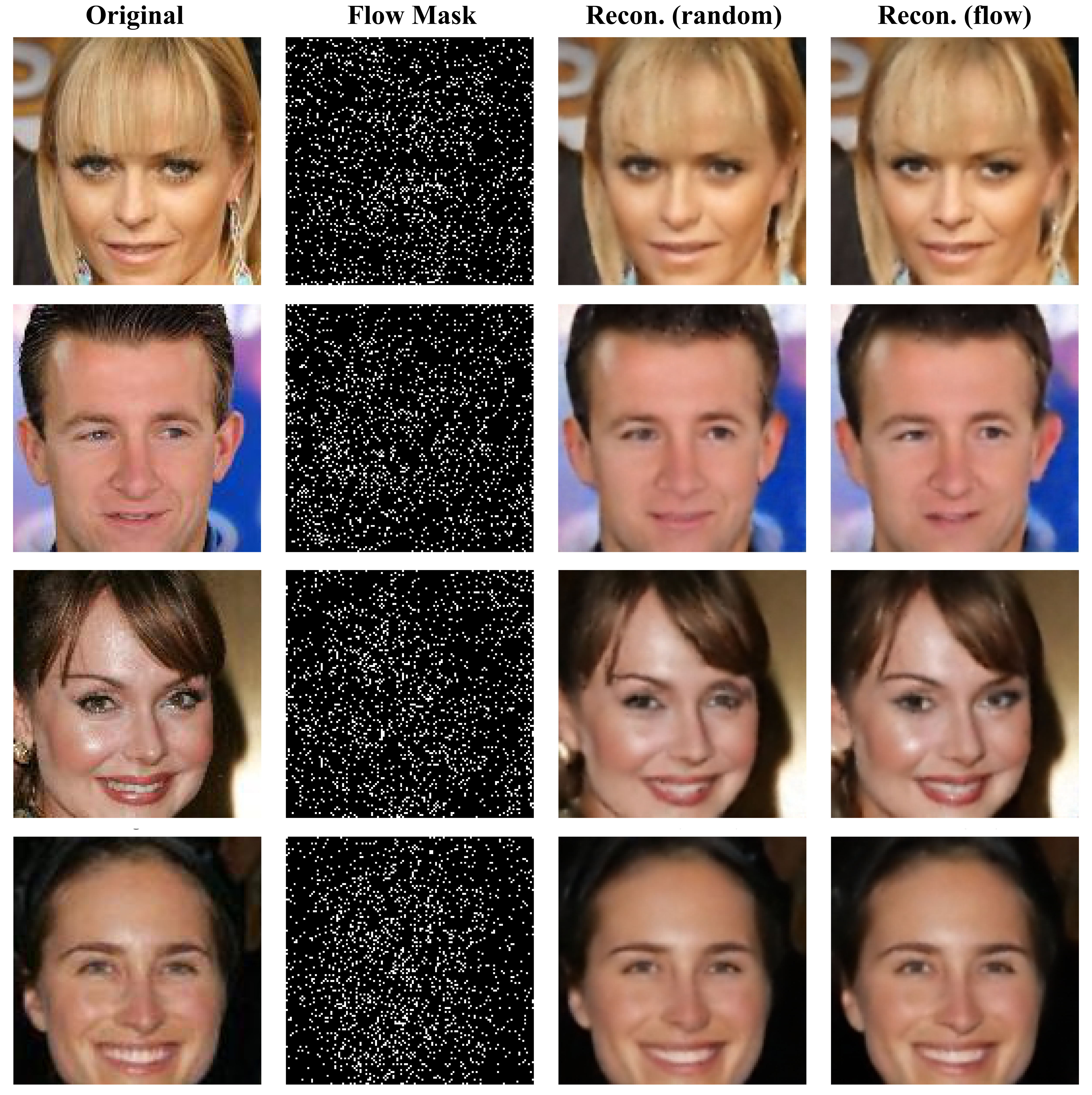}
    \caption{Test CelebA samples, corresponding flow-generated sampling masks (sampling rate $10\%$), and the U-Net reconstructions (random and the flow masks).}
    \label{fig:fig_5}
\end{figure}

Figure~\ref{fig:fig_5} presents selected representative images from the test set, accompanied by the corresponding flow-generated sampling masks (sampling rate $10\%$) and the associated U-Net-based reconstructions.

\subsection{MRI acceleration}
Table~\ref{tab:tab_1} presents the mean structural similarity index~(SSIM) quality scores of MRI slices for several baselines. The image reconstructed by applying the inverse Fourier transform to the zero-filled subsampled $k$-space data~(ZF) is used as the reference baseline.

\begin{table}[tb]
\centering
{\fontsize{9pt}{11pt}\selectfont
\caption{MRI acceleration performance on the fastMRI knee dataset for different reconstruction methods.}
\label{tab:tab_1}
}
\vspace{0.1in}
\begin{tabular}{p{2.1cm}cc}
\toprule
\raggedright \textbf{Method} &
\textbf{$4\times$ (SSIM)} &
\textbf{$8\times$ (SSIM)} \\
\midrule
\textbf{ZF}            & 0.7184 & 0.5665 \\
\textbf{LOUPE}         & 0.8952 & 0.5670 \\
\textbf{DPS}           & 0.9013 & 0.5680 \\
\textbf{ADS}           & \textbf{0.9126} & --     \\
\textbf{A-DPS}         & -- & 0.5820 \\
\midrule
\textbf{Ours}          & 0.8584 & \textbf{0.7163} \\
\bottomrule
\end{tabular}
\end{table}

At an acceleration factor of 4, the flow model in combination with MoDL yields images with lower SSIM values compared to the other evaluated models, although the resulting image quality remains substantially superior to that of the baseline. The performance of the proposed framework differs radically at $8\times$ acceleration. The flow-based approach in the full regime confidently outperforms all other algorithms, as follows from the highest SSIM score. 

\begin{figure*}[tb]
    \centering
    \includegraphics[width=0.85\linewidth]{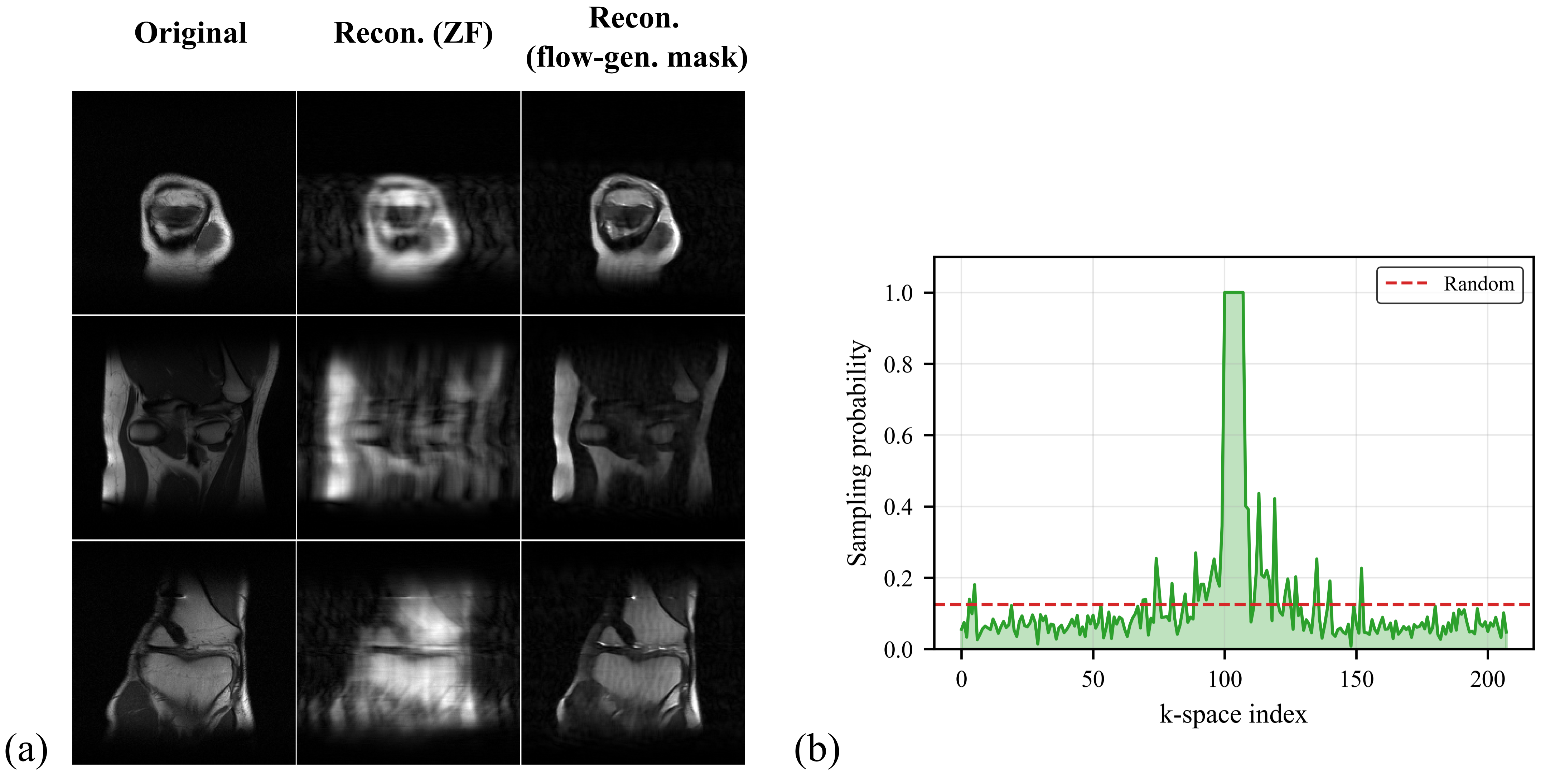}
    \caption{(a) Examples of $8\times$ accelerated MRI reconstruction (from the knee fastMRI dataset): original (\textbf{left}); zero-filled reconstruction (\textbf{middle}); Ours (\textbf{right}); (b) average subsampling mask generated across the test set.}
    \label{fig:fig_6}
\end{figure*}

In Figure~\ref{fig:fig_6}(a), MRI images from the test set reconstructed from 8 times accelerated measurements using our method are shown in comparison with the zero-filled baseline. The~distribution of sampled pixels in $k$-space, obtained by averaging the generated masks over the test set, is shown in Figure~\ref{fig:fig_6}(b). 
It can be seen that the model gives preference to frequencies in the vicinity of the hard-coded LF band. The sampling probability decreases as one moves away from this band into the high-frequency range. At the same time, noticeable peaks appear in the tails of the distribution, for example, near the frequencies with indices 80, 130, and 150. Notably, the distribution of the flow-generated masks differs from that of ADS masks (see Appendix A.4(b) in~\cite{nolan_active_2024}); in particular, the latter exhibits a smoother profile.

Finally, we measured the inference time for our framework to output a sampling scheme given a conditioning LF measurement. This consideration is essential for establishing the clinical utility and overall validity of the proposed framework in real-world practice. On our NVIDIA GeForce RTX 4070, we could achieve the average mask generation rate of $\approx$~130~ms (with 0.99~quantile of $\approx$~275~ms) for $4\times$~acceleration. For $8\times$~acceleration, our method performed similarly:  $\approx$~127~ms of average generation time with 0.99~quantile of $\approx$~236~ms. For reference, conventional acquisition times for individual $k$-space lines in MRI are on the order of 500~ms to 2000~ms, or even longer in some protocols~\cite{jung_spin_2013}. The reported acquisition time per line for ADS method was 3040~ms~\cite{nolan_active_2024}, whereas our algorithm achieves an acquisition time of approximately 4~ms per sampling line. For the majority of clinical settings, the associated computational overhead would be negligible, thereby enabling real-time execution of the proposed framework.

These results clearly demonstrate the rationale for the potential use of the proposed task-aware flow-based generation in clinical practice.

\section{Conclusion}
\label{conc}
This proof-of-concept study presents a flow-based, task-aware generative framework that optimizes sampling strategies for compressed sensing. By learning physical models that prioritize task-relevant data, the model generates sampling patterns that significantly outperform random allocation in both image classification and reconstruction. Furthermore, our framework achieves performance that is comparable to or surpasses current state-of-the-art methods for both image and MRI reconstruction.

The presented results stem from a reformulated Flow Matching training paradigm designed for settings lacking ground-truth data, which prevents standard linear coupling between noise and data domains. Instead of relying on such couplings, the model learns to generate intermediate "soft masks" at each time step, directly optimizing performance for specific downstream compressed sensing tasks.

The results validate the proposed framework, demonstrating that its continuous sampling policy generation and smooth manifold parametrization make it highly adaptable to various inverse problems. Future applications include the generation of point spread functions for image deblurring, producing object-level segmentation masks, and addressing other related downstream tasks.

\section*{Acknowledgments}
The authors would like to thank Dr. Hemant Kumar Aggarwal for his valuable assistance in debugging the MoDL-based MRI reconstruction framework used in this work.

%Bibliography
\bibliographystyle{unsrt}  
\bibliography{references}

\newpage
\appendix
\label{append}
\onecolumn

\section{Ablation studies on the parameters of the Sigmoid gate.}
\label{append_a}

\begin{figure}[h]
    \centering
    \includegraphics[width=0.6\linewidth]{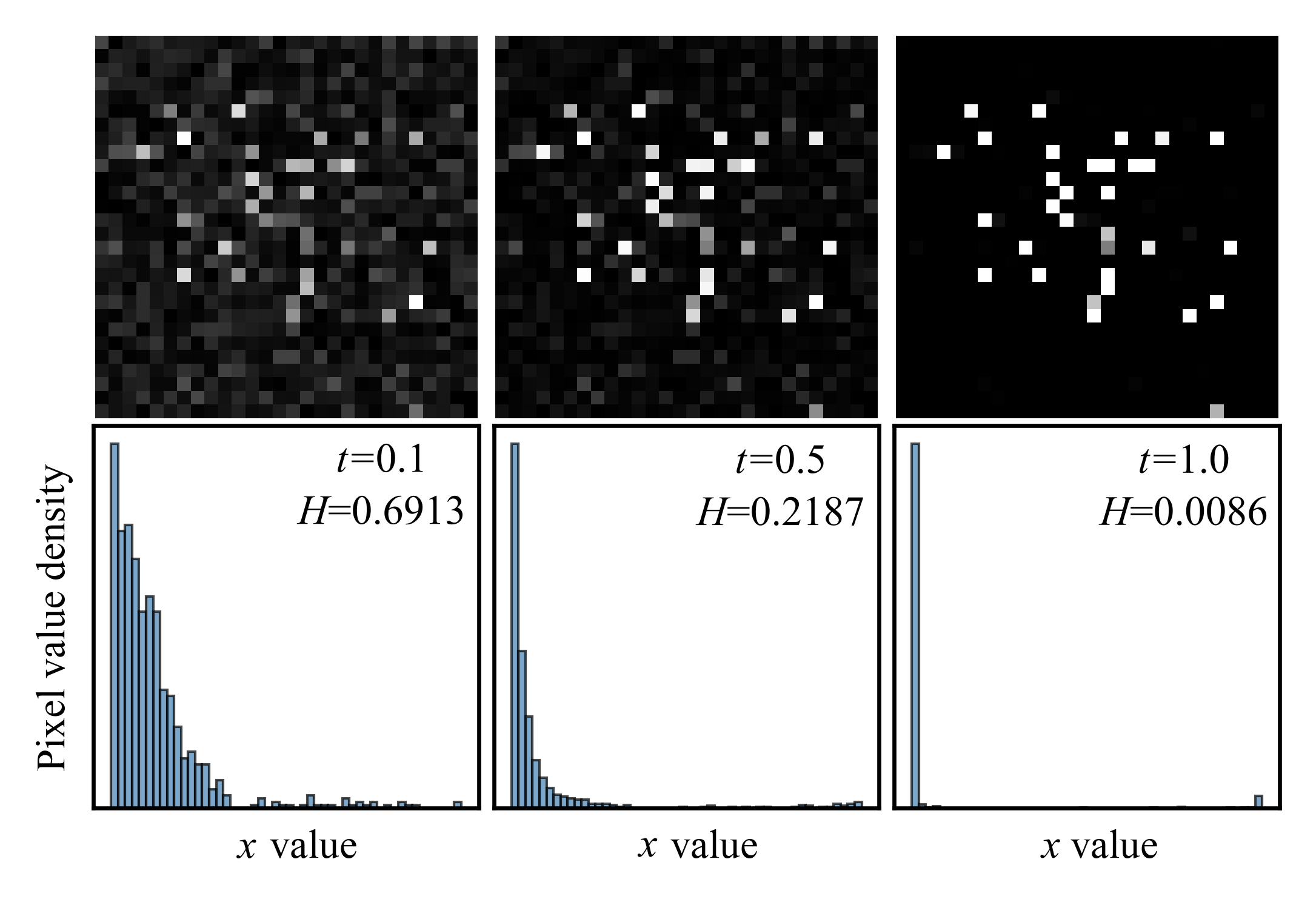}
    \caption{Examples of the effect of the function $\mathcal{S}(\cdot)$ on input $x$ at different time steps: transformed soft masks $g$ (\textbf{top~row}); distribution of the pixel values of the corresponding soft masks (\textbf{bottom~row}). Here, $H$ denotes the binary entropy averaged over all pixels.}
    \label{fig:fig_7}
\end{figure}

As it was outlined in Section~\ref{methods}, $t$ effectively regularizes the entropy of the marginal distribution $p_t$ (soft mask), similar to the temperature parameter in the Gumbel-softmax estimators of sampling schemes~\cite{huijben_review_2022}. Figure~\ref{fig:fig_7} is provided to illustrate this effect.

To ensure an appropriate gradient scaling for training $v_\phi$ controlled by the steepness $s=at^b$ of the Sigmoid~gate~Eq.~(\ref{eq:eq_11}), we made a series of short training runs (10 epochs) for the task of MNIST classification to define the parameters~$a$~and~$b$ minimizing the total loss~Eq.~(\ref{eq:eq_10}). The resulting accuracies achieved at different sampling rates are reported in Table~\ref{tab:tab_2}.

\begin{table}[h]
\centering
{\fontsize{9pt}{11pt}\selectfont
\caption{Ablation study to define the best parametrization of the Sigmoid gate.}
\label{tab:tab_2}
}
\vspace{0.1in}
\begin{tabular}{p{2.1cm}ccccc}
\toprule
$a$ & $b$ & \textbf{Acc., 1\%} & \textbf{Acc., 4\%} & \textbf{Acc., 8\%} \\
\midrule
1 & 1 & 40.50 & 71.02 & 75.52 \\
10 & 1 & 42.62 & 75.77 & 82.15 \\
20 & 1 & 44.50 & 75.64 & 82.96 \\
30 & 1 & 44.86 & \textbf{76.61} & \textbf{83.90} \\
40 & 1 & \textbf{45.03} & 75.16 & 82.30 \\
30 & 2 & 43.69 & 74.88 & 81.70 \\
\bottomrule
\end{tabular}
\end{table}

To obtain the estimated ground truth distribution $\tilde{p}_1$ to calculate the FM loss term in~Eq.~(\ref{eq:eq_10}), we used the DPS framework of Huijben~et~al.~\cite{huijben_deep_2019}, which provides a fixed probability density in the sampling domain.

\section{Pseudocode of the Flow-based Compressed Sensing Sampling Policy Learning algorithm}
\label{append_b}

\begin{algorithm}[h]
\caption{Flow-based CS Sampling Policy Learning}
\label{alg:tagm_algorithm}
\begin{algorithmic}[1]
\REQUIRE Pre-trained frozen encoder $\mathcal{R}$, flow model $v_\phi$, task model $f_\theta$ (can be frozen/pre-trained or not)

\FOR{$\text{iter} = 1$ to epochs}
    \STATE Sample batches $\{y^B, z^B\} \sim \mathcal{D}_{\text{train}}$
    
    \STATE \textcolor{blue}{// Add noise and subsample}
    \STATE $\tilde{y} \gets y + \epsilon, \quad \epsilon \sim \mathcal{N}(0, I)$
    \STATE Create random masks $\hat{A}$ with $\|\hat{A}\|_0 = M$ sampling pixels
    \STATE $\tilde{y}^{\text{sub}} \gets \tilde{y} \odot \hat{A}$
    
    \STATE \textcolor{blue}{// Flow matching: predict the subsampling mask logits}
    \STATE Get conditioning features: $l \gets \mathcal{R}(\tilde{y}^{\text{sub}})$
    \STATE Get samples from the surrogate GT distribution $\tilde{p}_1$: $x^{\star} \gets \text{Top-}K(\mathcal{F}(M))$ \quad \textcolor{gray}{// $\mathcal{F}$ is Maximum Likelihood Estimation (MLE) sampler or DPS-based sampler, $K=M$ for the hard thresholding}
    \STATE Sample $x_0^B \sim \mathcal{N}(0, I), \quad t^B \sim \mathcal{U}(0, 1)$
    \STATE Define linear FM path and target velocity: $\tilde{x}_t = (1-t)x_0 + t x^{\star},\qquad
    u^{\star} = x^{\star} - x_0$
    \STATE $v_t^ \gets v_\phi(\tilde{x}_t, t, l)$ \quad \textcolor{gray}{// Predict velocity field}
    \STATE $x_t \gets x_0 + v_t \cdot t$ \quad \textcolor{gray}{// Generate marginal mask logits}
    
    \STATE \textcolor{blue}{// Task-aware sampling}
    \STATE Evaluate the percentile thresholds: $\tau \gets \mathrm{Quantile}(x_t, 1-M/N)$
    \STATE $A^{\text{soft}} \gets \mathcal{S}\!\left( s(t) \cdot (x_t - \tau) \right)$ \quad \textcolor{gray}{// Differentiable sigmoid gate}
    \STATE $\tilde{y}^{\text{soft}} \gets \tilde{y} \odot A^{\text{soft}}$ \quad \textcolor{gray}{// Apply soft sampling}
    
    \STATE \textcolor{blue}{// Task prediction and optimization}
    \STATE $\hat{z} \gets f_\theta(\tilde{y}^{\text{soft}})$ \quad \textcolor{gray}{// Task prediction}
    \STATE Calculate the total loss $\mathcal{L}_{\text{total}}(z^B, \hat{z}^B)$
    \STATE Update $\phi$ and $\theta$ (if joint training) via $\nabla_{\theta, \phi} \mathcal{L}_{\text{total}}$
    
\ENDFOR

\STATE \textbf{return} Trained model $v_\phi$ and $f_\theta$ (if joint training)
\end{algorithmic}
\end{algorithm}

\section{Model architectures and implementation details.}
\label{append_c}

This appendix contains tables that provide details on the adopted flow matching and task model architectures for the different experiments. 

The reconstruction model for MNIST was trained for 200 epochs with a batch size of 64 using the Adam optimizer with a learning rate of $1 \times 10^{-4}$ and gradient clipping with a maximum norm of 1.0. Training inputs were corrupted with Gaussian noise (STD=0.02) and random subsampling; the model was optimized using $L^2$-loss to recover clean images. The $v_\phi$ decoder utilizes LeakyReLU activations (0.2 negative slope) and Batch Normalization for stability. Detailed layer configurations for the flow and classification networks are provided in Tables~\ref{tab:tab_3} and \ref{tab:tab_4}.

In the experiment with CS of CelebA images the architecture of $\mathcal{R}$ and $v_\phi$ remained the same as in the experiment with MNIST images (see Table~\ref{tab:tab_3}). Both networks were trained for 10 epochs with a batch size of 32. The subsampling rate varied from 4.88\% to 48.83\%, analogous to the original experiment conducted by Nolan et al. $\mathcal{R}$ and $v_\phi$ were trained with a learning rate of $1~\times~10^{-4}$. Both the image encoding and flow models (task loss) were trained to minimize the $L^1$-loss.

MoDL is an unrolled optimization framework. It formulates reconstruction (Eq.~(\ref{eq:eq_1})) as a regularized inverse problem:
\begin{equation}
    x_{\mathrm{rec}} = \arg\min_{\mathbf{x}}
    \underbrace{\|A(x) - z\|_2^2}_{\text{data consistency}}\;+\;
    \lambda\,\underbrace{\|x-\mathcal{D}_{w}(x)\|_2^2}_{\text{regularization}};
    \label{eq:eq_14}
\end{equation}
and alternates between two modules: (1)~a data-consistency update, which enforces agreement with the acquired undersampled $k$-space measurements using the explicit MRI forward model (including the sampling mask $A$), and (2)~a CNN-based denoiser $\mathcal{D}_{w}$, which acts as a learned regularizer that removes aliasing and undersampling artifacts. These two operations are repeated for a fixed number of iterations, each iteration forming one “unrolled” stage of the network. From an operational perspective, MoDL is advantageous as the forward model and data-consistency step are explicitly differentiable.

Our MoDL implementation inherits most of the aspects of the original network presented by Aggarwal~et~al.~\cite{aggarwal_modl_2018}. The organization of the residual CNN denoiser $\mathcal{D}_{w}$ is given in Table~\ref{tab:tab_4}. The data consistency step solves the linear system:
\begin{equation}
    (\mathcal{F}^H \mathcal{F} + \lambda I)x = \mathcal{F}^H k + \lambda \mathcal{D}_{w}(x),
    \label{eq:eq_15}
\end{equation}
where $\mathcal{F}$ represents the forward operator (Fast Fourier Transform (FFT) followed by masking), $\mathcal{F}^H$ is the adjoint (masking followed by inverse FFT), $I$ is the identity operator, $k$ is the measured subsampled $k$-space data, and $\lambda$ is a learnable regularization parameter. The solution is computed using the differentiable conjugate gradient (CG) algorithm, iterating 8 times in our experiments. As in the original paper, we first pre-trained a MoDL network for a single unrolled iteration for 5 epochs, and then fine-tuned it for 10 unrolled iterations for 10 epochs to minimize the $L^2$ loss. 

For the training, random Gaussian subsampling masks were generated via selecting a specified percentage of central frequencies (8\% for the $4\times$ case and 4\% for the $8\times$ case), and the rest of the subsampling lines were uniformly distributed. The resulting subsampled $k$-space undergoes channel-specific noise corruption: Gaussian noise is added independently to real and imaginary components, with standard deviations computed as a percentage of each channel's magnitude standard deviation. The flow model was trained with the frozen MoDL task model, employing a batch size of 4 across 1,000 iterations, and the fixed learning~rate~of~$1~\times~10^{-4}$ with the task objective to minimize the $L^2$-loss and (1 - SSIM)-score between the original and reconstructed images.

\begin{table}[h]
\centering
\footnotesize
\caption{The $\mathcal{R}$ network employed in Section~\ref{methods_class_mnist}--\ref{methods_recon_mnist}. Spatial resolutions are given in pixels.}
\label{tab:tab_3}
\vspace{0.1in}
\begin{tabular}{lccc}
\toprule
\textbf{Stage} & \textbf{Resolution} & \textbf{Channels} & \textbf{Operation} \\
\midrule
Input          & $28 \times 28$ & 1   & -- \\
\midrule
Encoder 1      & $28 \times 28$ & 1 $\rightarrow$ 32   & Conv + MaxPool \\
Encoder 2      & $14 \times 14$ & 32 $\rightarrow$ 64  & Conv + MaxPool \\
Encoder 3      & $7 \times 7$   & 64 $\rightarrow$ 128 & Conv + MaxPool \\
Bottleneck     & $3 \times 3$   & 128 $\rightarrow$ 256 & Conv \\
\midrule
Decoder 1      & $7 \times 7$   & 256 $\rightarrow$ 128 & UpConv + Skip(Enc3) \\
Decoder 2      & $14 \times 14$ & 128 $\rightarrow$ 64  & UpConv + Skip(Enc2) \\
Decoder 3      & $28 \times 28$ & 64 $\rightarrow$ 32   & UpConv + Skip(Enc1) \\
Output         & $28 \times 28$ & 32 $\rightarrow$ 1    & Conv (final) \\
\bottomrule
\end{tabular}
\vspace{0.1in}
\end{table}

\begin{table}[h]
\centering
\footnotesize
\caption{The classification network from Section~\ref{methods_class_mnist}. FC stands for Fully Connected layer.}
\label{tab:tab_4}
\vspace{0.1in}
\begin{tabular}{lccc}
\toprule
\textbf{Layer} & \textbf{Input Dim} & \textbf{Output Dim} & \textbf{Activation / Notes} \\
\midrule
FC1     & 784 & 256 & Leaky ReLU, Dropout 30\% \\
FC2     & 256 & 128 & Leaky ReLU, Dropout 30\% \\
FC3     & 128 & 128 & Leaky ReLU, Dropout 30\% \\
FC4     & 128 & 10  & Softmax \\
\bottomrule
\end{tabular}
\vspace{0.1in}
\end{table}

\begin{table}[h]
\centering
\footnotesize
\caption{The residual CNN denoiser $\mathcal{D}_{w}$ for the MoDL framework. The denoiser operates on two-channel inputs (real and imaginary parts).}
\label{tab:tab_5}
\vspace{0.1in}
\begin{tabular}{lccc}
\toprule
\textbf{Block} & \textbf{Input Channels} & \textbf{Output Channels} & \textbf{Operations} \\
\midrule
Conv Block 1 & 2   & 32 & Conv2D $\rightarrow$ BN $\rightarrow$ ReLU \\
Conv Block 2 & 32  & 32 & Conv2D $\rightarrow$ BN $\rightarrow$ ReLU \\
Conv Block 3 & 32  & 32 & Conv2D $\rightarrow$ BN $\rightarrow$ ReLU \\
Conv Block 4 & 32  & 32 & Conv2D $\rightarrow$ BN $\rightarrow$ ReLU \\
\midrule
\multicolumn{4}{c}{Residual connection: input added to final output} \\
\bottomrule
\end{tabular}
\vspace{0.1in}
\end{table}

\begin{table}[h]
\centering
\footnotesize
\caption{The flow matching model from Section~\ref{methods_mri_acc}.}
\label{tab:tab_6}
\vspace{0.1in}
\begin{tabular}{lcc}
\toprule
\textbf{Component} & \textbf{Output Dim.} & \textbf{Details} \\
\midrule
Frozen CNN encoder & -- & Extracts multiscale 2D features (from $\mathcal{D}_{w}$) \\
Global Avg. Pooling & $F$ & Pools all CNN feature maps into a vector \\
Sinusoidal time embedding & 64 & Standard positional encoding of $t$ \\
Concatenated input & $d_x + F + 64$ & $x_t$ + CNN features + time embedding \\
\midrule
MLP Hidden Layer 1 & 512 & Linear $\rightarrow$ LayerNorm $\rightarrow$ SiLU \\
MLP Hidden Layer 2 & 512 & Linear $\rightarrow$ LayerNorm $\rightarrow$ SiLU \\
MLP Hidden Layer 3 & 512 & Linear $\rightarrow$ LayerNorm $\rightarrow$ SiLU \\
MLP Hidden Layer 4 & 512 & Linear $\rightarrow$ LayerNorm $\rightarrow$ SiLU \\
Output Layer & 208 & Linear \\
\bottomrule
\end{tabular}
\vspace{0.1in}
\end{table}

\end{document}